\documentclass[runningheads]{llncs}
\usepackage{graphicx}
\usepackage[hyphens]{url}
\usepackage{hyperref}
\hypersetup{breaklinks=true}
\usepackage[absolute,overlay]{textpos}
\graphicspath{{./Figures/}}

\begin{document}
\title{Autonomy, Authenticity, Authorship and Intention in computer generated art}
\titlerunning{AAAI in computer generated art}
%
\author{Jon McCormack\orcidID{0000-0001-6328-5064} \and
 Toby Gifford\orcidID{0000-0002-9902-3362} \and
 Patrick Hutchings\orcidID{0000-0001-8680-6969}}
\authorrunning{J. McCormack et al.}
%
\institute{SensiLab, Faculty of Information Technology, Monash University, Caulfield East, Victoria, Australia 
\email{Jon.McCormack@monash.edu}\\
\url{https://sensilab.monash.edu}}

\maketitle              
%
\begin{textblock*}{\textwidth}(4.2cm,24cm) 
  \fbox{%
    \parbox{\textwidth}{\small%
        Preprint of: J.~McCormack, T.~Gifford and P.~Hutchings, `Autonomy, Authenticity, Authorship and Intention in computer generated art', in 
        \textit{EvoMUSART 2019: 8th International Conference on Computational Intelligence in Music, Sound, Art and Design}, April 2019, Leipzig, Germany, 2019
          } }
\end{textblock*}

\begin{abstract}
This paper examines five key questions surrounding computer generated art. Driven by the recent public auction of a work of ``AI Art'' we selectively summarise many decades of research and commentary around topics of autonomy, authenticity, authorship and intention in computer generated art, and use this research to answer contemporary questions often asked about art made by computers that concern these topics. We additionally reflect on whether current techniques in deep learning and Generative Adversarial Networks significantly change the answers provided by many decades of prior research.

\end{abstract}
\section{Introduction: Belamy's Revenge}
\label{s:introduction}
In October 2018, AI Art made headlines around the world when a ``work of art created by an algorithm'' was sold at auction by Christie's for US\$432,500 -- more than 40 times the value estimated before the auction \cite{Christies2018}. The work, titled \textit{Portrait of Edmond Belamy} was one of ``a group of portraits of the fictional Belamy family''%
\footnote{The name is derived from the French interpretation of ``Goodfellow'': \emph{Bel ami}} created by the Paris-based collective \textit{Obvious}.

The three members of Obvious had backgrounds in Machine Learning, Business and Economics. They had no established or serious history as artists. Their reasoning for producing the works was to create artworks ``in a very accessible way (portraits framed that look like something you can find in a museum)'' with the expectation of giving ``a view of what is possible with these algorithms.''\cite{Obvious2018}

The works' production involved the use of Generative Adversarial Networks (GANs), a technique developed by Ian Goodfellow and colleagues at the University of Montreal in 2014 \cite{Goodfellow2014}. 
It turned out that Obvious had largely relied on code written by a 19 year old open source developer, Robbie Barrat, who did not receive credit for the work, nor any remuneration from the sale (and who in turn, relied on code and ideas developed by AI researchers such as Goodfellow and companies like Google). An online arts commentary site, Artnet, summarised it thus: ``Obvious\ldots was handsomely rewarded for an idea that was neither very original nor very interesting'' \cite{Rea2018}.

\begin{figure}
\begin{center}
\includegraphics[scale=0.5]{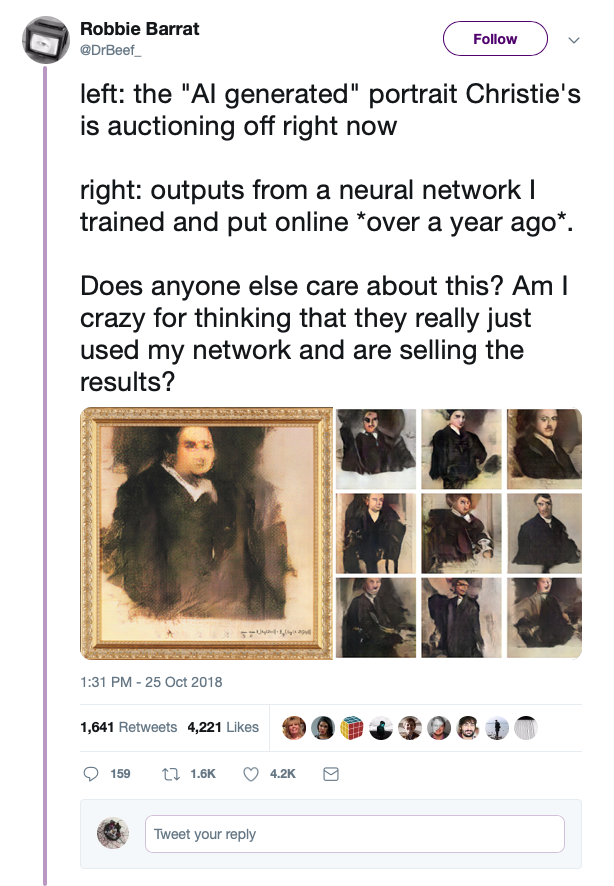}
\caption{Tweet by Robbie Barrat following the auction of \textit{Portrait of Edmond Belamy}} \label{f:BarratTweet}
\end{center}
\end{figure}

During the auction, Barrat complained on twitter about the situation (Fig. \ref{f:BarratTweet}):
\begin{quote}
    Does anyone else care about this? Am I crazy for thinking that they really just used my network and are selling the results?
\end{quote}

Following the sale of \textit{Portrait of Edmond Belamy} to an undisclosed buyer, social media erupted with 
amazement, skepticism, and in some instances, outrage. Many publicly expressed concern that artists who had spent many years working with AI and art, and specifically GANs, had been overlooked; that their work represented far better examples of ``AI Art'' than Obvious' crude and derivative efforts. Ethical concerns were raised about Obvious being the ``authors'' of a work generated using code written by a third party (Barrat, whose code was based on Soumith Chintala's implementation of DCGANs \cite{Radford2015}), even though Obvious were legally entitled to use it under the open source license under which it was released.

The sale of this work, and the attention it has subsequently brought to what was, until then, a relatively obscure area of research and creative production, raises many questions. Of course those with knowledge of history understand that creative applications of AI techniques date back to the earliest days of the use of the term ``Artificial Intelligence'' and even predate it.\footnote{Ada Lovelace is famously known as one of the first people to record ideas about computer creativity \cite{Menabrea1842}} The EvoMUSART workshop and conference series has been running since 2003 and has presented numerous examples of ``art made by an algorithm''.\footnote{A great resource is the EvoMUSART Index, available at: \url{http://evomusart-index.dei.uc.pt}} The pioneering work of artists such as Michael Knoll, Georg Nees, Frieder Nake, Lilian Schwartz and Harold Cohen is widely acknowledged in both technical and artistic surveys of computer art and dates back to the early 1960s. Indeed, Cohen -- a trained artist of some renown -- spent most of his working life in the pursuit of AI Art, yet his works typically did not sell for even 1\% of the price paid for \textit{Portrait of Edmond Belamy}.

The sale and subsequent reaction to the work resurrects venerable questions regarding autonomy, authorship, authenticity, and intention in computer generated art. It should come as no surprise that, despite public reaction to the sale of \textit{Portrait of Edmond Belamy}, these are well explored issues (see e.g.~\cite{Burnham1968a,Burnham1969,Reichardt1971b,Davis1973,Leavitt1976,Dietrich1983,McCorduck1990,Ohear1995,Cohen1995,BownM09,Boden2010,DBLP:conf/evoW/McCormack13,McCormack2014,dinvernoMcCormack2015}). It is worth revisiting them for two reasons: firstly to encapsulate more than half a century of discussion around computer created art, and secondly, to see if recent advances in Machine Learning techniques, such as Deep Learning and GANs, change any of the well established research and commentary in this area.

\subsection{Questions Raised}
\label{ss:questions}

In this paper we review and comment on work in this area drawn from philosophy, psychology, cognitive science and computational creativity research, and use it to answer the following five key questions:

\begin{enumerate}
    \item Who is the author of a work of art made by an AI algorithm?
    \item What attribution or rights should be given to developers of an ``art'' generating algorithm?
    \item To what extent is any AI artwork relying on mimicry of human creative artefacts (rather than being independently creative)?
    \item Is art made by ``Deep Learning'' techniques such as GANs different than art made by other generative algorithms?
    \item Do artworks made by AI represent a new ``kind'' of art?
\end{enumerate}

There are many additional interesting questions that could be considered in relation to AI art generally, and specifically in the case of \textit{Portrait of Edmond Belamy}. For example, questions surrounding the extraordinary price paid for the work, the motivations of the creators, the quality of the work as an exemplar of the genre of AI Art, and so on. However answering these questions needs to draw on broader fields of research, such as market and social dynamics in the art world and auction system, human behavioural science and art criticism, and is beyond the scope of what is covered in this paper. A good introduction to these issues can be found in \cite{Thornton2009}, for example.

\subsection{AI Art}
\label{ssaiArt}
Throughout this paper we use the term ``AI Art'' in it's most obvious meaning: art that is made using \emph{any} kind of AI technique. There is no real consensus on what is included or excluded from ``AI Art''. For example, evolutionary techniques are often considered a branch of AI, but traditionally, art made using evolutionary techniques is not referred to as AI Art. Many practitioners who adopt this term use deep learning techniques, such as GANs, Pix2Pix (Conditional Adversarial Networks), etc. For any cases in the paper where we need more specificity, we will refer to specific techniques.

\section{Autonomy, Authenticity, Authorship and Intention (AAAI)}
\label{s:aaai}
Art is a human enterprise, defined by experience, of both artist and audience. Demarcations of the continuous actions and/or material manipulations of an artist into art-objects (be they paintings, sculptures, poems, scripts, scores or performances) take place in myriad ways, though commonly through the intentional declaration of an artist that a particular artefact is to be considered a finished work. Whether the result of years of meticulous crafting, such as Michelangelo's \textit{David}, or readymades such as Duchamp's \textit{Fountain}, the ``trait inherent in the work of the artist [is] the necessity of sincerity; the necessity that he shall not fake or compromise'' \cite{dewey1934art}. The artist's intention, then, is critical, and intertwined with issues a work's authenticity. Even more  fundamental is the autonomy of the artist and their authorship when considering the aesthetic value of an artefact. Thus when considering work generated, in part or whole, by an artificial intelligence, the extent to which these properties can be ascribed to the machine is brought into question.

\subsection{Autonomy}
\label{ss:autonomy}

One of the main attractions of working with generative computational systems is their capacity for agency and autonomy. Systems that can surprise and delight their authors in what they produce are undoubtedly an important motivation for working with computer generated creative systems \cite{McCormack2017}.

Boden provides an extensive examination of autonomy in computer art \cite[Chapter 9]{Boden2010}. She distinguishes between two different kinds of autonomy in non-technological contexts and their parallels in computational art:

\begin{itemize}
    \item physical autonomy such as that exhibited in homeostatic biological systems, and
    \item mental/intentional autonomy typified by human free-will.
\end{itemize}

Boden describes how these different kinds of autonomy are at play in generative art. Many concepts used in generative art, such as emergence, agency, autopoesis, homoeostasis and adaptation are underpinned by \emph{self-organisation}; something Boden views as synonymous with a specific kind of autonomy where, ``the system's independence is especially strong: it is not merely self-controlled, but also self-generating'' \cite[p. 180]{Boden2010}, with the ``self'' in self-organisation referring to the impersonal components of the system, not the intentional, mental self.

Boden's second form of autonomy is inherently tied to human freedom: something lacking in the autonomy of AI art systems that use an emergent, ``bottom up'' philosophy (such as Artificial Life or Evolutionary based systems). This intentional autonomy requires concepts from the Theory of Mind, such as intention, desire and belief. There is nothing in principle that dismisses the possibility of artificial creative systems possessing this intentional autonomy, and indeed, this is something extensively explored by Artificial Intelligence research since its inception. If this class of autonomy is dependent on free-will, then any artificially autonomous creative software would potentially need to make the choice to make art, not be explicitly programmed to do so. 

\subsubsection{Autonomy of Deep Learning Systems} \label{sss:autonomy}
With this general distinction in mind, it is worth considering how deep learning techniques, such as GANs fit into these different categories of autonomy. A GAN was used by Obvious to create \emph{Portrait of Edmond Belamy}. GANs use two competing neural networks, a \emph{generator} and a \emph{discriminator}. The generator learns to map from a latent space to some (supplied) data distribution, while the discriminator attempts to discriminate between the distribution produced by the generator and the real distribution.\footnote{This is conceptually similar to co-evolutionary strategies, such as creator/critic systems \cite{Todd1998}, used at least since the 1990s.}

Perceptions of autonomy in GANs rest on their ability to generate convincing ``fakes'' (convincing to the discriminator at least). They have autonomy in the sense of being able to synthesise datasets that mimic the latent space of their training data -- in many current artistic works this training data comes from digitised works of other artists, often those recognised in the cannon of classical or modernist art. Certainly they have a very limited autonomy in choosing what statistical model might best represent the latent space and in the datasets generated. In visual examples, this is often akin to structured ``mashups'': images with similar overall appearances to parts of images from the training set, recalling the visually similar, but algorithmically different technique of \emph{morphing} \cite{Beier1992}, developed in the early 1990s.

In contrast, human artists do not learn to create art exclusively from prior examples.  They can be inspired by experience of nature, sounds, relationships, discussions and feelings that a GAN is never exposed to and cannot cognitively process as humans do. The training corpus is typically highly curated and minuscule in comparison to human experience.  This distinction suggests GANism is more a process of mimicry than intelligence.  As flowers can utilise Dodsonian mimicry to lure pollinators by generating similar sensory signals \cite{dodson}, current GAN techniques lure human interest by mimicking valued aesthetic properties of prior artistic artefacts without being a direct copy \cite{tan2017artgan,he2018chipgan,zhang2017multi,elgammal2017can}.

At a technical level, GANs are a technique for finding a statistical distribution of latent dimensions of a training set. Hence their capacity for autonomy is limited, certainly no more that prior machine learning or generative systems. Even in their perceived autonomy as image creators, their ability to act autonomously is limited within a very tight statistical framework that is derived from their training data. There is no evidence to suggest that they posses any of Boden's second sense of intentional autonomy.

Thus any claims we can make about the autonomy of a GAN-based software system's autonomy are limited. Certainly  many different generative systems with equal or greater autonomy exist (in the non-intentional sense). While a claim such as, ``an AI created this artwork'' might be literally true, there is little more autonomy or agency that can be attributed to such an act than would be to a situation where ``a word processor created a letter'', for example.

\subsubsection{Summary}
\label{sss:summary}
Artistic generative systems illuminate the complex relationships between different kinds of autonomy. Artists and researchers have worked with such systems because they value their ability to exploit autonomous (in the self-organising, homeostatic sense) processes, while often (incorrectly) ascribing autonomy in the intentional sense to them. Nonetheless, autonomous generative processes result in more aesthetically challenging and interesting artworks over what can be achieved when the computer is used as a mere tool or ``slave''. They open the possibility for types of expression not possible using other human tools. However, that does not give them the intentional autonomy characterised by human creativity.

\subsection{Authorship}
\label{ss:authorship}
Issues of agency and autonomy also raise another important concept challenged by generative art: authorship. Who is the author of an artwork that is generated by a computer program? A computer program that can change itself in response to external stimuli has the potential to learn and adapt (autonomy in the self-organising sense). Hence it can do things that the programmer never anticipated or explicitly designed into the software, including the potential to act creatively, even if this ability was not part of its original programming (see e.g.~\cite{Keane:1996}). Thus while we may describe the programmer as the author of the program, it seems that our commonsense notion of authorship should extend to recognising the program itself as at least a partial author of the artwork in such circumstances. Commonsense, however, may be culturally determined and historically fraught; and the notion of authorship has a vexed and contested history.

\subsubsection{Authorship and Individuality}
To talk of an author presupposes the notion of an \textit{individual}; a notion that only developed into something close to its current form in western philosophy in the 17\textsuperscript{th} century, and even today has different forms in many eastern cultures. Take Thomas Hobbes' 1651 definition of a person: ``He whose words or actions are considered either as his own or as representing the words or actions of any other thing to whom they are attributed, whether truly or by fiction'' \cite{hobbes1651}. Hobbes distinguishes one whose words are considered their own as a \textit{natural person} and one whose words are considered as representing another an \textit{artificial person}. Only natural persons does Hobbes deign `authors'. 

Hobbes' definition of a person appears to preclude female persons within the very first word. This highlights the relativism of notions such as person-hood and authority to the ambient culture. The question of whether or not a computer could ever be considered an ``author'' may be tractable only to cultural argument, rather than relating to either objective truth or universal human experience.

An instructive case study is that of Albert Namatjira, one of the first Indigenous Australian artists to adopt a European landscape style in his paintings. So enamoured were the colonial powers of this ``civilised'' Aborigine (who even met the Queen) that Albert was honoured as the first Indigenous Australian to be granted citizenship of Australia in 1957. Here we see perhaps the person-author relationship operating inversely to the received wisdom: his artistic authorship (in the culturally approved style) propelled him from non-entity into person-hood.

When Namatjira died he bequeathed his copyright (possession of which was one perk of being considered a natural person) to his family, and subsequently his paintings have become enormously valuable. Unfortunately his family did not enjoy person-hood, and so the bequest was managed by the state, which chose (without consultation) to sell all rights to a private investor for a pittance. The family has been fighting ever since for the copyright to be returned to their community, and have only recently succeeded in re-obtaining it.

This story is by no means unique, nor restricted to the visual arts. The music world is replete with examples of traditional musics being co-opted by colonisers, recorded and copyrighted, the original performers subsequently persecuted for continuing to perform it. The notion of authorship here is simply an expression of power. As times progress, with womens' suffrage, Indigenous citizenship, and increasing egalitarianism so too the granting of authorship has expanded. Perhaps when AIs are granted citizenship they too will receive legal recognition as authors, although the first such citizen robot -- Sophia -- has yet to produce any creative works. More likely, as with Albert Namatjira, the causality will flow the other way; when AIs start autonomously creating good art (in the culturally accepted style), then robots will be granted status as people.

\subsubsection{Authorship and Romanticism}
Our current understanding of authorship is rooted in the Romantic movement. Whilst contemporary culture has abandoned many Romantic ideas, including that of the lone genius, the concept of authorship has remained relatively fixed, probably due to the development of copyright and intellectual property laws in the late Enlightenment and early Romantic period, and the ensuing substantial vested interests in its perpetuation. Jaszi \cite{jaszi1991toward} suggests ``it is not coincidental that precisely this period saw the articulation of many doctrinal structures that dominate copyright today. In fact British and American copyright present myriad reflections of the Romantic conception of `authorship' ''. For example, in a mid 20\textsuperscript{th} century legal case an artist superimposed two images from the \textit{Wizard of Oz} and attempted to claim copyright on the result (preempting copyright disputes that would follow over mashups and remixes). The judge rejected that application saying ``we do not consider a picture created by superimposing one copyrighted photographic image on another to be original'' \cite{jaszi1991toward}. Jaszi argues this decision has no real basis in the legal arguments made, but that:
\begin{quote}
this decision does make sense, however, when viewed in light of the Romantic ``authorship'' construct, with its implicit recognition of a hierarchy of artistic productions. In that hierarchy, art contains greater value if it results from true imagination rather than \textit{mere application}, particularly if its creator draws inspiration directly from nature. \cite{jaszi1991toward}
\end{quote}
Echoes of this debate continue in the computational creativity field, surfacing in the contest between critics of \textit{mere generation} \cite{veale2015scoffing} and computational media artists for whom computer generated artefacts (be they complete, partial, or co-constructed) form an integral part of their practice \cite{eigenfeldt2016flexible}.

\subsubsection{Legal and Moral Ownership}\label{sss:legal}
The concept of authorship, as affected by non-anthropocentric intelligence and cognitive labour is also of interest in legal discussions around intellectual property, and has been for several decades.  In 1997 Andrew Wu \cite{wu1997video} wrote on issues of copyright surrounding computer-generated works.  He perceived a gap between the objective of copyright and the idea of copyrighting works produced by an intelligent computer.
\begin{quote}
... the basic problem presented by intelligent computers; awarding copyright to the one who is the author--in the sense of being the originator or intellectual inventor of a work--does not further the objective of stimulating future creativity. \cite{wu1997video}
\end{quote}

Copyright is intended to encourage discovery and creation by reducing the fear that creators might have of others benefiting from their work instead of the creators themselves.  Currently, computers don't have the economic needs, emotional concerns or desires that drive copyright legislation, and so the incentive to provide copyright to software systems is greatly reduced.  At the same time there is an awareness that software can have a significant role in the creation of artefacts that are deemed to be creative works.

While legal interpretations will differ between countries, it is clear that artificial intelligence presents unique challenges for the concept of copyright and was a point of discussion for legal scholars in the last century.

In 1984 the US Copyright Office listed a computer program named ``Racter'' as an author of \textit{The Policeman's Beard is Half Constructed} but not a claimant of the copyright \cite{racter}.  The distinction between claimant and author is interesting because it contrasts the modern construct of companies as economically driven legal entities that can claim ownership over physical and virtual goods, but are not deemed to be authors of any artefact themselves.

Today the same questions are still being raised around authorship for copyright.  In arguing for patent protection for computer generated works (CGWs), Ryan Abbott suggests that legal changes to copyright may have been slowed by a lack of commercial need:

\begin{quote}
Given these technological advances, one would be forgiven for asking — where are the CGWs? Why are there not routinely lawsuits over CGWs? How have countries managed without legal standards for CGWs?
It may be that the creative AI revolution has yet to arrive. CGWs may be few and far between, or lack commercial value. \cite{abbott2017artificial}
\end{quote}

Now that artworks made with generative systems have resulted in significant commercial gains there may be increased interest in establishing legal standards around them.  There may also be a shift in the dynamics of AI art as a movement and community.  Until now, artists have both relied on and contributed to the distribution of open-source code.  The benefits of community and reputation building by sharing open-source code have outweighed potential economic returns for AI art, but have left little basis for claims of legal ownership of these systems.

Without legal ownership, authors maintain a moral right to partial ownership of the process.  This ownership can be expressed in the naming of a technique or algorithm, or honest and accurate attribution.  Software and datasets, especially those utilising creative works of other individuals, are attributable contributions to machine learning-based systems.  

\subsubsection{Summary}
Ascribing authorship of a creative work is only meaningful in a particular cultural and legal context. In a contemporary Western context, authorship of AI generated art will be shared between the artist, the developer of the underlying algorithms, and possibly any number of contributors to the training data whose style is being abstracted. Increasingly the algorithms themselves may be accepted as partial authors, both morally and legally, in step with their increasing perception as autonomous intelligent agents. However, the level of intelligence of existing AI systems falls short of its perception in current popular imagination. As such, greater attention should be focused on accurate and inclusive attribution amongst the constellation of human actors. Accurate attribution not only benefits these authors, but helps establish the authenticity of work produced with AI systems.

\subsection{Authenticity}
The authenticity of computer art often provokes quite binary responses. A number of theorists, critics and philosophers rule out the idea of computer art entirely being authentic because the art was made by a machine, not a person \cite{Ohear1995}.

A common criticism of works employing algorithms is that they are \emph{generic}, and such genericism implies that they lack authenticity. As was the case with \textit{Portrait of Edmond Belamy}, the human creators relied on software derived from a common pool of algorithms and code, and the Obvious team played only a very minor role in actually even writing any software \cite{verge}. The majority of algorithms used for computer art have originated from the sciences, not the arts, so understandably they evoke scepticism when introduced as key components in an artistic process \cite{Parikka2008}.

These and similar criticisms can be characterised by three related points:

\begin{enumerate}
    \item That works made with the same or similar algorithms -- even by different artists -- possess a certain generic and repetitive character;
    \item Artists do not fully understand, or are misleading about, the process of creation, in that the algorithms do not exhibit the characteristics or outcomes that the artists' claim the processes used represent;
    \item Works exclude artistic possibilities due to their dependence on a generic technical process that arose in an unrelated context.
\end{enumerate}

Addressing the first point, it is typically the case that similar algorithms produce similar results. Repeatability and determinism are fundamental utilities of computation after all. And if the algorithm is actually the ``artist'' then it makes sense that the same algorithm would produce works that are the same or with highly similar characteristics.\footnote{It is worth noting that artistic styles or idioms share similar characteristics too.}
There is nothing inherently so generic in creative software in general -- a tool like Photoshop is theoretically capable of creating any style or kind of image. This is because most of the \emph{creative agency} rests with the user of the software, not the software itself \cite{BownM09}.  Machine learning researcher, Francois Chollet has referred to art made by GANs as \emph{GANism}\footnote{\url{https://twitter.com/fchollet/status/885378870848901120}}, implying the algorithm is responsible for the style more than the artists who use the software. As deep learning software becomes increasingly complex and difficult to understand, authorship and creative agency shifts towards the algorithm itself. However, in the case of reinforcement learning, training data plays a crucial role in determining what the learning system does. When the training data is authored by others, they implicitly make a contribution, akin to the way that mashups, collage or sampling have contributed to artworks previously (see Section \ref{ss:authorship}). 

Algorithmically generated art systems emphasise process as the primary mechanism of artistic responsibility. A generic or copied process will produce generic work. In considering generative systems in an artistic context, the ``Art'' is in the construction of process. If that process is derivative or memetic then the work itself will likely share similar properties to others, making any claims to artistic integrity minimal.\\

\subsubsection{Truth about Process}
\label{sss:disclosure}
Outside of technical research circles, terms such as ``Artificial Intelligence'' inherently carry assumed meaning, and the value of this fact has not escaped numerous entrepreneurs and marketing departments who claim their work has been ``created by an AI.'' 
People will naively find anything associated with non-human ``intelligence'' interesting, simply because we aren't used to seeing it in the wild. But because the algorithms and techniques are largely opaque to non-experts, it is natural -- but incorrect -- to assume that an artificial intelligence mirrors a human intelligence (or at least some major aspects of human intelligence). Such views are reinforced by the anthropomorphic interfaces often used for human-AI interaction, such as realistic human voices, scripted personalities, etc. 

Recent analysis of public perception of AI indicates that non-experts are more generally optimistic about the possibilities of AI than experts \cite{Manikonda2018}. In one survey 40\% of respondents ``think AI is close to fully developed'' \cite{Gaines-Ross2016}. Such surveys reinforce the problem of dissonance between public perceptions and reality regarding technological capability.

\begin{quote}
    We don't ascribe artistic integrity to someone who produces art in indefinitely many styles on request \cite[p. 187]{Boden2010}
\end{quote}

Any full disclosure of artistic process in AI art needs to appreciate the authenticity of the generative process as fundamental to the artwork. Current AI Art, including that made using GANs, are \emph{not} intelligent or creative in the way that human artists are intelligent and creative. In our view, artists and researchers who work with such techniques have a responsibility to acknowledge this when their work is presented and to correct any misunderstandings where possible.\\

\subsection{Intention}
\label{ss:intention}

Another well discussed issue in relation to computer generated art is that of \emph{intention}. How can a software program be considered an ``artist'' if it lacks the independent intention to make art or to \emph{be} an artist? Software artists are employed like slaves: born into servitude of their creators to relentlessly and unquestionably be art production machines. We would never force a human intelligence to only be concerned with producing art, nor would we consider an artist just an art-production machine, yet we have no difficulty saying that artificial intelligence must serve this role.

As discussed in Section \ref{ss:autonomy}, the kind of autonomy typified by free-will that human artists possess includes the ability to decide not to make art or to be an artist. Software programmed to generate art has no choice but to generate; it lacks ``the intentional stance'' \cite{Dennett1987} of mental properties and states that might lead to a consideration of making art in the first place. So in addition to lacking autonomy, AI Art also lacks intention. Such conditions are often glossed over in everyday descriptions, where anthropomorphism and personalising are commonplace.

The terminology surrounding algorithmically generated art systems has shifted in recent years.  While deep neural networks share many mathematical properties with Markov models and can employ evolutionary algorithm techniques for training, systems utilising deep learning have quickly gained the label of ``artificial intelligence'' while existing techniques are typically labelled ``generative''.  Intelligence has a much stronger association with intent than does generation, and the shift in terminology is likely affecting the perception of intention in these systems.

\section{Questions Answered}
\label{s:answers}

Having now examined the issues of autonomy, authenticity, authorship and intention, we are now able to answer the questions posed in Section \ref{ss:questions}.

\subsection{Who is the author of a work of art made by an AI algorithm?}

 The creator of the software and person who trained and modified parameters to produce the work can both be considered authors. In addition to these contributors, artists whose works feature prominently in training data can also be considered authors.  There has been precedent cases of software systems being considered legal authors, but AI systems are not broadly accepted as authors by artistic or general public communities. See Section \ref{ss:authorship}.

\subsection{What attribution or rights should be given to developers of an ``art'' generating algorithm?}

Authors have a responsibility to accurately represent the process used to generate a work, including the labour of both machines and other people.  The use of open-source software can negate or reduce legal responsibilities of disclosure, but the moral right to integrity of all authors should be maintained. See Section \ref{sss:disclosure}.

\subsection{To what extent is any AI artwork relying on mimicry of human creative artefacts?}

AI systems that are trained to extract features from curated data-sets constructed of contents produced by people are relying on mimicry of artefacts rather than autonomously searching for novel means of expression.  This holds true for current, popular AI art systems using machine learning. See Section \ref{sss:autonomy}. 

\subsection{Is art made by ``Deep Learning'' techniques such as GANs different than art made by other algorithms?}

There are no significant new aspects introduced in the process or artefact of many GAN produced artworks compared to other established machine learning systems for art generation.  Currently, there is a difference in the way GANs are presented by media, auction houses and system designers: as \textit{artificially intelligent systems} that is likely affecting the perception of GAN art.  As this difference is grounded more in terminology and marketing than intrinsic properties of the technique, history suggests it is not likely to sustain.  In her 1983 paper clarifying the terminology of ``Generative Systems'' and ``Copy Art'' Sonia Sheridan stated:

\begin{quote}
    Despite our efforts, ``Copy Art'' emerged to exploit a single system for a marketable, recognizable art product. In our commercial society and in the context of conditions prevailing through the seventies, such a development was inevitable, as was the struggle to maintain the character and integrity of the Generative Systems approach. \cite{sheridan1983generative}
\end{quote}

See Sections \ref{ssaiArt} and \ref{ss:intention}.

\subsection{Do current artworks made by AI represent a new ``kind'' of art?}

Probably not in any major way. At least no more than any other kind of computer generated art. And let us not forget that computer generated art is now more than 50 years old. Certainly there are some stylistic differences between GAN Art and other types of computer generated art, just as there are many different painting styles. However, the idea of a computer autonomously synthesising images isn't new. Nor are the conceptual ideas or process presented in contemporary AI Art new. It might be argued that the technical methodology is new, even if it is largely developed by non-artists for non-artistic purposes. However, differences in technical methods rarely constitute a new kind of art on their own -- cultural and social factors typically play a far greater role.

\bigskip
\subsubsection{Summary} 
We hope this paper has helped clarify and reiterate some of the large body of research around the status of computer generated art, particularly those works that claim to be authored, in full or in part, by non-human intelligence. We leave it to the reader to make judgement on the merits or otherwise of the sale of \textit{Portrait of Edmond Belamy} and the claims made by Obvious and Christies.

\section{Acknowledgements}
This research was support by Australian Research Council grants DP160100166 and FT170100033.

%
%
%
\bibliographystyle{splncs03}

\end{document}